%% file: paper.tex
\PassOptionsToPackage{table}{xcolor}
\documentclass[]{fairtum}
\pdfoutput=1

\usepackage{amsmath,amssymb}
\usepackage{natbib}
\usepackage{siunitx}
\usepackage{environ}
\usepackage{fancyvrb}
\usepackage{adjustbox}
\usepackage{wrapfig}
\usepackage{flafter}
\usepackage{pgfplots}

\usetikzlibrary{arrows.meta,positioning,calc,fit}
\pgfplotsset{compat=1.18}

\graphicspath{{./}{assets/}{figures/}}

\fancypagestyle{firstheader}{%
  \fancyhf{}
  \fancyfoot[C]{\thepage}

}

\newenvironment{ack}{\section*{Acknowledgments}}{}

\title{Relational Learning as World Modeling}

\newcommand{\github}{\raisebox{-1.5pt}{\includegraphics[height=1em]{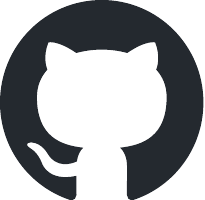}}}
\newcommand{\huggingface}{\raisebox{-1.5pt}{\includegraphics[height=1em]{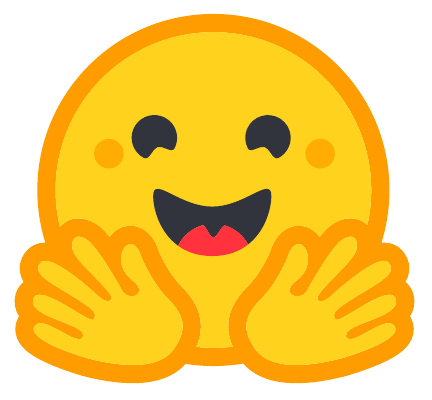}}}

\author{Zheyu Zhang \& Gjergji Kasneci }

\affiliation{Technical University of Munich\\
Munich Center for Machine Learning (MCML)\\
{\href{https://github.com/oooranz/TabWM}{\github{} Github}\quad
\href{https://huggingface.co/oooranz/TabWM}{\huggingface{} Model}}}

\makeatletter
\let\abstract\relax

\NewEnviron{abstract}{%
  \global\edef\abstractlist{{%
    \noexpand\color{metafg}\unexpanded\expandafter{\BODY}}}}
\makeatother

\begin{abstract}
Relational learning typically predicts from a database state, but decision making asks a different question: how will that state change under an action? In operational systems, the answer depends on system-specific dynamics that may not be visible in the database itself. We view each such system as a relational ``world'' and ask whether a model can infer how it behaves from observed executions. We introduce \textbf{TabWM} (\textbf{Tab}ular \textbf{W}orld \textbf{M}odel) and \textsc{WorldRel}, a generator of executable relational worlds for pretraining. Given executions from an unseen world, TabWM uses them as context to predict the sparse structured edits caused by a new action without parameter updates. Across 512 held-out worlds, complete-edit F1 rises from 0.016 with no observed executions to 0.685 with 64, where a correct edit must match its operation, relational location, and value. Because these predictions are explicit relational edits, they can be applied recursively to imagine candidate futures and compare actions before they are taken. Broader search over these predicted futures improves the realized return of selected action sequences. Together, these results suggest that relational dynamics can be learned in context, broadening relational learning from modeling system state to modeling how that state changes under action.
\end{abstract}

\metadata[Correspondence]{\email{zheyu.zhang@tum.de}}

\begin{document}
\maketitle
\thispagestyle{firstheader}


\begin{figure}[!ht]
    \vspace{-5pt}
    \centering
    \includegraphics[width=.85\linewidth]{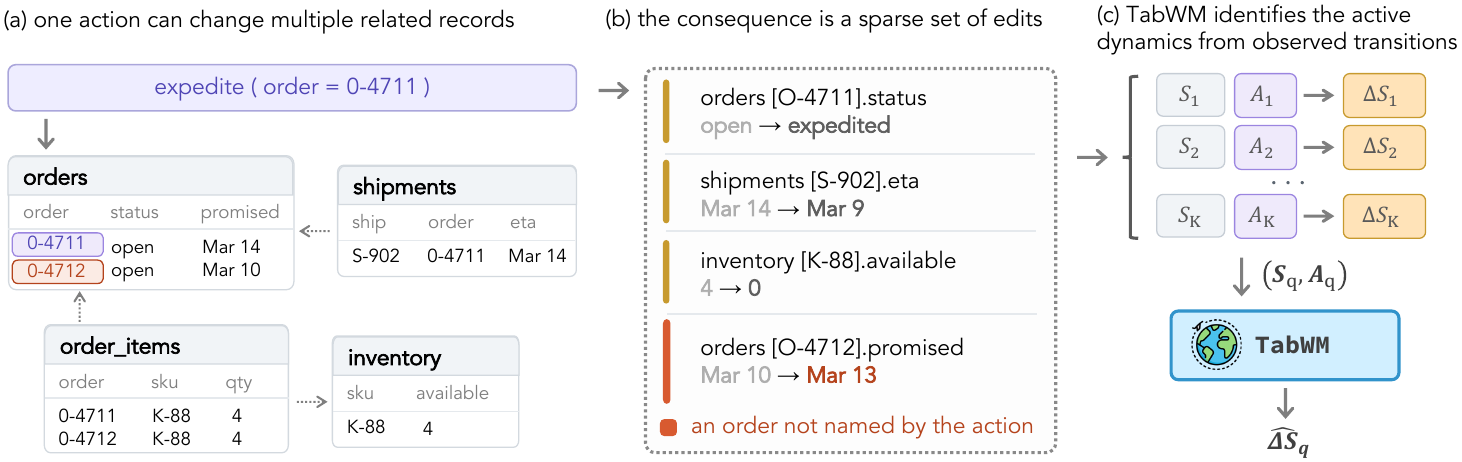}
    \caption{Relational learning as world modeling. \textbf{(a)} An action can affect both referenced and linked records. \textbf{(b)} Its consequences form a sparse edit set. \textbf{(c)} TabWM uses observed executions from the same world to predict edits for a new action without parameter updates.}
    \label{fig:overview}
\end{figure}
\vspace{-15pt}

\section{Introduction}
\label{sec:introduction}

Relational foundation models are rapidly expanding what a single pretrained model can learn from databases. Recent models transfer across unseen schemas and tasks~\citep{ranjan2026relational,hudovernik2026kumorfm2}, scale through synthetic relational pretraining~\citep{kothapalli2026plurel}, and adapt to new prediction problems from context~\citep{wang2026relational,ranjan2026largescale}. Yet these advances largely share the same learning interface, using a relational database state $S$ to predict a target $y$. This formulation treats the database primarily as a dataset. In many applications, however, the database is also the observable state of a running system, shaped by users, software, and operational processes~\citep{klein2025position,klein2026statistically}. Choosing an action requires asking how it would change that state. This calls for an action-conditioned transition model from $(S_t,A_t)$ to $S_{t+1}$, the core abstraction of world modeling that allows a model to imagine how a system evolves under candidate actions~\citep{ha2018recurrent,ding2025understanding,hafner2025mastering}. We ask whether relational learning can make this shift from prediction to \emph{world modeling}.

Consider an order that is predicted to arrive late. Knowing that it will be late does not tell us what will change if we expedite it. In Figure~\ref{fig:overview}a, expediting O-4711 updates its order and shipment, consumes the remaining inventory, and pushes back O-4712 although the action does not name it. Similar what-if questions arise in prescriptive process monitoring and business-process simulation~\citep{bozorgi2023prescriptive,camargo2020automated,camargo2022learning,chapelacampa2025simod}. We represent these direct and propagated changes as a sparse set of structured edits $\Delta S_t$ (Figure~\ref{fig:overview}b). Each edit specifies an operation, its relational location, and a value when needed. Applying the edits updates the affected records without regenerating unchanged data. The action specifies what is done, but the system dynamics determine where its effects propagate and what values result.

These dynamics depend on operational rules that may not be apparent from the database records~\citep{klein2025position,klein2026statistically}. We call a relational system with a particular schema, set of actions, and transition mechanism a \emph{world}. Its executions are observable, but we assume its mechanism is hidden. In a controlled experiment across $200$ pairs of worlds, executing the same action from identical relational states produces different structured edits in $65\%$ of probes. The current state and action therefore leave the consequence ambiguous across worlds. Past executions provide evidence about how actions behave in the current world. Intuitively, \emph{states show what a system looks like, while transitions show how it behaves}.

Learning from these executions is often approached by discovering or calibrating a simulator for a particular process~\citep{camargo2020automated,camargo2022learning,chapelacampa2025simod}. In-context system identification offers another route. A model trained across a class of dynamical systems can infer the behavior of a new one from its observed history~\citep{forgione2023classmodels,rufolo2024enhanced}. We ask whether the same idea can be extended to relational systems. Instead of fitting a separate model for every new world, a single pretrained model would use observed executions as context to predict the outcome of a new action in that world (Figure~\ref{fig:overview}c).

We realize this idea with \textbf{TabWM} (\textbf{Tab}ular \textbf{W}orld \textbf{M}odel), trained across executable relational worlds generated by \textsc{WorldRel}. \textsc{WorldRel} extends the PluRel relational prior~\citep{kothapalli2026plurel} with executable actions and world-specific transition mechanisms. Within a task, the observed executions and query share the same dynamics, while those dynamics vary across tasks. This extends prior-data fitted learning~\citep{muller2022pfn,wang2026relational} from static relational prediction to action-conditioned transitions. TabWM preserves relational structure together with which actions produced which observed consequences. It then predicts the consequence of a new action as a sparse set of edits that specify what changes, where it changes, and what value is written. At test time, the model adapts to a new world through observed executions without updating its parameters.

Across $512$ held-out worlds, complete-edit F1 rises from $0.016$ with no behavioral context to $0.685$ with $64$ observed executions, where a correct edit must match its operation, relational location, and value. Context remains useful when action mappings, schemas, and mechanism compositions change, and the same pattern persists when relational structure and values are derived from real databases. A single pretrained model can therefore improve its predictions from experience in a new world without updating its parameters.

A world model is useful not only for predicting what happens next, but for deciding what to do next. Because TabWM predicts explicit edits, it can roll candidate actions forward into future relational states and compare the resulting trajectories through model-based planning~\citep{hafner2019planet,chua2018pets}. At a fixed three-step horizon, expanding the candidate pool from $16$ to $64$ improves the selected return in $30$ of $40$ held-out worlds. This gives TabWM two ways to improve at test time. More observed executions improve consequence prediction, while broader search over predicted futures improves action selection.

\section{Problem Setup}
\label{sec:problem}

\subsection{Relational Worlds}
\label{sec:relational_worlds}

We study a relational system whose execution rules are hidden from the learner. Its structure and dynamics define a world
\begin{equation}
w=(\Sigma_w,\mathcal A_w,\rho_w,T_w),
\label{eq:relational-world}
\end{equation}
where $\Sigma_w$ specifies the tables, columns, and relational links, $\mathcal A_w$ the available actions, $\rho_w$ the distribution of initial execution states, and $T_w$ the transition mechanism. These components remain fixed while the database state changes through execution.

An action $A_t$ specifies its type and arguments, which can reference records or supply values~\citep{pasula2007symbolic,martinez2016relational}. The database is the learner's observation of the system, not its complete execution state. We write $S_t=\operatorname{Obs}(X_t)$, where $X_t$ also contains any unobserved information used during execution. Starting from $X_0\sim\rho_w$, executing an action produces
\begin{equation}
\Delta S_t\sim T_w(\cdot\mid X_t,A_t).
\label{eq:execution}
\end{equation}
The consequence $\Delta S_t$ is a variable-size set of edits that update numeric or categorical attributes, insert records, or delete them. Each edit specifies an operation, its relational target, and any required values or record contents. Numeric updates specify resulting values, not increments. The target records the changes caused during action execution, excluding separately generated external changes. It is empty when execution produces no edits.

\subsection{Predicting from Behavioral Context}
\label{sec:behavioral_context}

The learner receives a behavioral context $\mathcal C_w^K=\{(S_i,A_i,\Delta S_i)\}_{i=1}^{K}$ of $K$ observed transitions and a query $(S_q,A_q)$ from the same world. The demonstrations include observed consequences, while the query consequence $\Delta S_q$ is withheld. The learner predicts
\begin{equation}
\widehat{\Delta S}_q=f_\theta(\mathcal C_w^K,S_q,A_q).
\label{eq:behavioral-inference}
\end{equation}
Demonstrations and queries share dynamics but come from separate episodes, each starting from a fresh initial state $X_0\sim\rho_w$. Context therefore provides evidence across executions, not the query's own history. The parameters $\theta$ are learned across worlds and held fixed for an unseen test world~\citep{forgione2023classmodels,wang2026relational}. Adaptation occurs through the observed transitions, without access to executable rules or complete execution states. We first test why this evidence is needed by holding the query state and action fixed while changing the execution rules.

\section{Same State, Different Consequences}
\label{sec:dott}

\begin{wrapfigure}{r}{.48\columnwidth}
    \centering
    \vspace{-0.7em}
    \includegraphics[width=\linewidth]{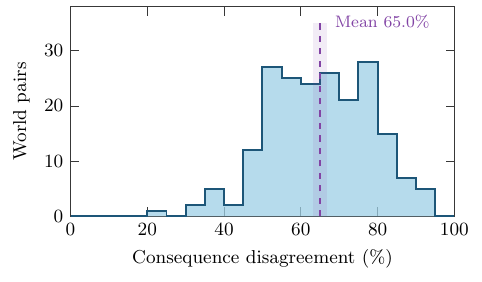}
    \caption{Edit-set disagreement under identical states and actions. The dashed line marks the mean across world pairs ($65.0\%$), with shading indicating bootstrap uncertainty in the mean.}
    \label{fig:dott_phenomenon}
    \vspace{-0.6cm}
\end{wrapfigure}

Similar database states can arise under different operational rules~\citep{klein2026statistically}. We test whether this ambiguity remains when the observed state and action are both identical.

We pair worlds that share a schema, initial-state distribution, observable action definitions, and set of transition mechanisms. We change only which mechanism each action invokes. We sample $200$ pairs without filtering and draw $64$ matched initial states per pair, checking that all database tables agree elementwise. Each probe executes the same valid action, including its arguments, in both worlds with the same random seed, giving $(S_0,A)\xrightarrow{w_0}\Delta S^{(0)}$ and $(S_0,A)\xrightarrow{w_1}\Delta S^{(1)}$.
We compare the complete edit sets, including each edit's operation, relational target, and required values or record contents. The observed states match in all $12{,}800$ paired probes, yet the edit sets differ in $65.0\%$ of them. Figure~\ref{fig:dott_phenomenon} shows how this disagreement rate varies across world pairs.

Observed transitions link actions to the changes they produced in the current world. They provide evidence about the execution rules that the state alone does not reveal. We next build a model that learns across worlds to use this evidence when predicting a new execution.

\section{\textbf{TabWM}: A \textbf{Tab}ular \textbf{W}orld \textbf{M}odel for Relational Systems}
\label{sec:method}

The ambiguity in \S\ref{sec:dott} turns consequence prediction into an identification problem: the same observed state and action can behave differently across worlds, so a model must use previous executions to infer which dynamics are active before predicting a new one. TabWM learns this inference across relational worlds while predicting the consequence itself rather than the hidden transition mechanism. Its training distribution varies dynamics across worlds, its encoder preserves the behavioral evidence within each observed execution, and its decoder converts that evidence into a sparse set of structured edits. Because these edits define the resulting relational state, the same model can also be applied recursively to imagine subsequent actions.

\subsection{Learning Across Relational Worlds}
\label{sec:world_pretraining}

Learning to identify a new world's dynamics from context requires training tasks in which behavior remains consistent within a world and varies across worlds. We construct such tasks with \textsc{WorldRel}, which extends the relational prior of \textsc{PluRel}~\citep{kothapalli2026plurel} with executable actions and world-specific transition mechanisms. Each sampled world fixes a relational schema, an action interface, an initial-state distribution, and a transition mechanism. Observed executions and the query are drawn from separate episodes of the same world, while worlds are resampled across tasks. \S~\ref{sec:worldrel_prior} describes this executable-world prior in detail.

The query consequence provides the prediction target, while the observed executions reveal which behavior applies in the current world. Following prior-data fitted learning~\citep{muller2022pfn}, training across these tasks teaches a single model to use behavioral context instead of relying on one fixed action--consequence mapping. This amortizes action-conditioned system identification across worlds~\citep{forgione2023classmodels}. At test time, TabWM adapts to an unseen world through its observed executions without updating its parameters.

\subsection{Encoding Relational Transitions}
\label{sec:transition_encoding}

Using executions as evidence requires preserving more than the database state alone. The representation must retain how records are linked, which action and consequence belong to the same execution, and how an observed value changed before and after that action. We initialize TabWM's relational encoder from the pretrained Relational Transformer~\citep{ranjan2026relational}, using weights learned from PluRel-generated relational databases~\citep{kothapalli2026plurel}. This provides an initial representation of relational structure, which TabWM adapts together with its transition-specific components to action-conditioned prediction.

For each execution, we collect an action-centered neighborhood by following primary- and foreign-key links from the records referenced by the action, subject to a cell budget. Cells retain their column identity, type, and value, while textual identifiers use deterministic hash embeddings rather than a pretrained language encoder. The action is represented as a pseudo-row linked to the records bound by its arguments, with scalar and categorical arguments represented as action cells.

We encode the correspondence within each execution. Demonstrated edits are attached to the records they modify. Role embeddings distinguish state, action, and consequence cells, while transition-index embeddings mark which elements belong to the same execution. For observed updates, shared table, record, and column identities align the before-state cell with its demonstrated edit. Together, these connections preserve relational structure, execution membership, and the before-and-after evidence needed to interpret the query. Packing the behavioral context and query gives
\begin{equation}
\mathcal U_q=\operatorname{Pack}(\mathcal C_w^K,S_q,A_q),
\qquad
H_q=\operatorname{Enc}_{\theta}(\mathcal U_q),
\label{eq:transition-encoding}
\end{equation}
where the query consequence is withheld and $H_q$ contains the resulting contextual representations.

\subsection{Decoding Action Consequences}
\label{sec:transition_prediction}

The encoder represents the query together with executions that may explain its behavior; the remaining problem is to predict a variable-size consequence. An action may produce several edits or none, so TabWM predicts its consequence as an unordered set. Following the set-prediction approach of DETR~\citep{carion2020end}, the decoder uses $M$ learned edit queries. Each query attends to $H_q$ and predicts an existence indicator $\hat z_j$, an operation $\hat o_j$, a relational target $\hat\ell_j$, and any required value or record contents $\hat v_j$. The active slots form
\begin{equation}
\widehat{\Delta S}_q
=
\left\{(\hat o_j,\hat\ell_j,\hat v_j)\mid 1\leq j\leq M,\ \hat z_j=1\right\}.
\label{eq:edit-slots}
\end{equation}
Pointers select affected cells, rows, or tables as required by the operation, while typed output heads produce categorical values and insertion contents.

Locating an edit does not yet determine the value to write. For a predicted numeric target, TabWM reads visible before-and-after pairs $(x_r,y_r)$ from previous executions, conditioned on the query action and target. Each pair links a value before an observed action to the value written by its consequence. A neural head produces a learned estimate $\hat y_{\mathrm{learned}}=\hat y_0+r_\theta$, while the same evidence is also used to infer transformations that can transfer to a query with a different starting value.

For a target whose current value is $x_q$, the decoder constructs additive and affine candidates,
\begin{equation}
h_{\mathrm{add}}=x_q+\sum_r\alpha_r(y_r-x_r),
\qquad
h_{\mathrm{aff}}=\bar y+\beta(x_q-\bar x),
\label{eq:numeric-candidates}
\end{equation}
where $\alpha_r$ are learned attention weights over relevant before-and-after pairs, $\bar x$ and $\bar y$ are their weighted means, and $\beta$ is an affine slope estimated from the selected evidence. The additive candidate transfers an observed increment, while the affine candidate estimates how the outcome varies with the starting value. Both transfer a transformation to the query rather than copying a demonstrated outcome.

Learned gates combine these candidates with the neural prediction,
\begin{equation}
\hat y
=
\hat y_{\mathrm{learned}}
+g_{\mathrm{add}}(h_{\mathrm{add}}-\hat y_{\mathrm{learned}})
+g_{\mathrm{aff}}(h_{\mathrm{aff}}-\hat y_{\mathrm{learned}}).
\label{eq:numeric-composition}
\end{equation}
The resulting value completes the predicted edit without changing its operation or target. This separation lets us isolate numeric transfer from relational effect localization in ~\S\ref{sec:numeric_reader_ablation}. Further details on evidence selection and candidate construction appear in \S\ref{sec:numeric_decoder_details}.

Minimum-cost bipartite matching associates predicted slots with reference edits during training~\citep{carion2020end}. Matched slots are supervised on their operations, targets, and contents, while unmatched slots learn to emit no edit.

Explicit edit predictions are also directly composable. After an imagined action, the predicted edits are applied to obtain the next relational state, which can then serve as the input to another action query:
\begin{equation}
\widehat{\Delta S}_t=f_\theta(\mathcal C_w^K,\widehat S_t,A_t),
\qquad
\widehat S_{t+1}=\operatorname{Apply}(\widehat S_t,\widehat{\Delta S}_t),
\qquad
\widehat S_0=S_0.
\label{eq:tabwm-rollout}
\end{equation}
The behavioral context remains fixed while the predicted state evolves along the action sequence. These rollouts produce the candidate futures used for action selection in \S\ref{sec:decisions}.

\section{Inferring Dynamics in Unseen Worlds}
\label{sec:identification}

If observed executions reveal how actions behave in a world, prediction should improve as more executions from that world enter context, without changing the model parameters. We test this by varying the amount of behavioral context available to TabWM.

\textbf{Evaluation.}
Each task provides $K$ observed state--action--consequence transitions and a separate query from a held-out \textsc{WorldRel} world. The context sweep uses $250$ worlds and three queries per world at each $K\in\{0,1,2,4,8,16,32,64\}$. Support F1 measures whether the predicted edits have the correct operations and relational locations, while Edit F1 additionally requires their written values to be correct. Value MAE measures numeric error once the correct location has been identified. The numeric tolerance and remaining scoring details are given in \S\ref{sec:evaluation_details}.

\begin{wrapfigure}{r}{0.42\textwidth}
\vspace{-3pt}
    \centering
    \includegraphics[width=\linewidth]{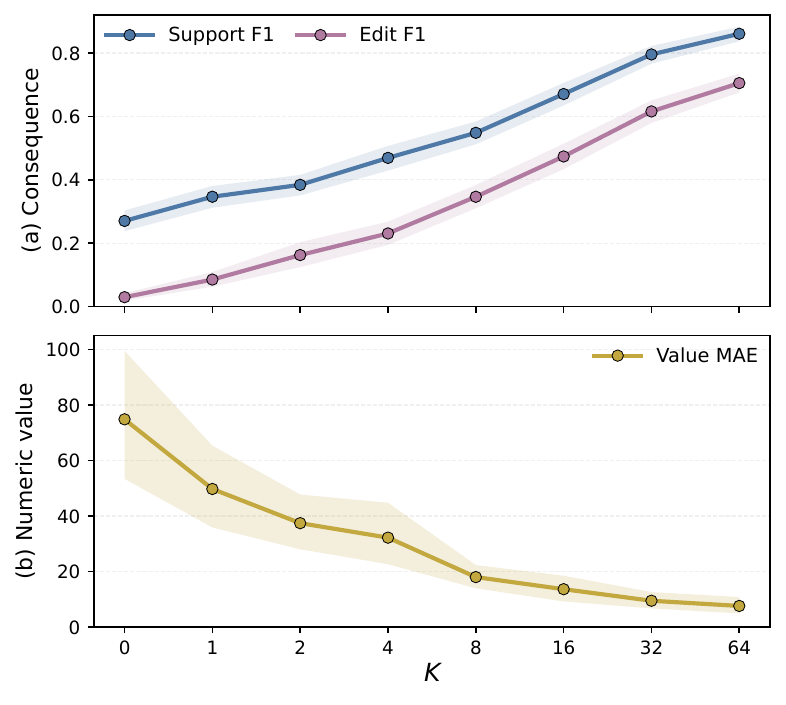}
    \vspace{-0.5cm}
    \caption{More observed behavior improves action--consequence prediction. \textbf{(a)} Support F1 and Edit F1 with increasing behavioral context. \textbf{(b)} Numeric MAE on correctly located edits. Shading shows 95\% world-bootstrap confidence intervals.}
    \label{fig:k_sweep}
    \vspace{-10pt}
\end{wrapfigure}

Observed executions improve both effect localization and complete-edit prediction, while numeric error on correctly located edits falls (Figure~\ref{fig:k_sweep}). Edit F1 reaches $0.705$ with $64$ executions, up from $0.029$ without behavioral context. Edit F1 still improves between $32$ and $64$ observations, so the context gain has not saturated by $32$ executions.

We next ask whether other model classes can exploit the same evidence. Flat-GBM and Flat-TabPFN reduce each task to fixed-width candidate-cell features and summaries of observed changes, then predict edits using gradient boosting~\citep{friedman2001greedy} or TabPFN~\citep{hollmann2025accurate}. Qwen3~\citep{yang2025qwen3} instead receives a textual serialization of the observed executions and query, testing whether a general-purpose LLM can infer the current dynamics from the same evidence.

Readable identifiers provide Qwen3 with an additional source of information because table, column, categorical, and action names may carry familiar linguistic cues. We therefore evaluate matched readable and opaque renderings of the same tasks. In the \emph{opaque} view, these identifiers are replaced consistently with arbitrary labels while preserving relational structure, row identities, values, types, and action arguments. The \emph{readable} view restores the original identifiers. TabWM uses deterministic hash embeddings, so readable strings do not provide pretrained linguistic semantics. All methods are evaluated on the same $60$ held-out worlds with one query per world at each $K\in\{0,4,16,64\}$, and Qwen3 and TabWM receive the same action-centred cell budget.

Under opaque naming, TabWM leads at every context size and reaches $0.687$ Edit F1 with $64$ observed executions, compared with $0.407$ for Qwen3-32B (Table~\ref{tab:behavior_identification}). The advantage is already visible in effect localization, where Support F1 is $0.887$ for TabWM and $0.753$ for Qwen3-32B under the same condition. Restoring readable names benefits Qwen3 substantially more than TabWM, but the complete-edit gap remains. Readable semantics therefore narrow the gap for the LLM, but do not account for TabWM's improvement from behavioral context.

\begin{table}[!ht]
\footnotesize
\caption{Complete-edit prediction (Edit F1, $\uparrow$) under opaque and readable naming cues. Readable names expose schema and action semantics. Best scores in each column are bold. $^{\dagger}$Flat baselines do not consume table or column names, so their inputs are unchanged between the two views.}
\label{tab:behavior_identification}
\begin{center}
\begin{adjustbox}{max width=.9\linewidth}
\begin{tabular}{l|c|cccc|cccc}
\toprule[0.8pt]
\multirow{2}{*}[-0.5ex]{\textbf{Model}} & \multirow{2}{*}[-0.5ex]{\textbf{\# Params}}
 & \multicolumn{4}{c|}{\textbf{Opaque names}} & \multicolumn{4}{c}{\textbf{Readable names}} \\
\cmidrule(lr){3-6} \cmidrule(lr){7-10}
 &
 & $K{=}0$ & $K{=}4$ & $K{=}16$ & $K{=}64$
 & $K{=}0$ & $K{=}4$ & $K{=}16$ & $K{=}64$ \\
\midrule
Flat-GBM$^{\dagger}$    & --            & $0.000$ & $0.005$ & $0.000$ & $0.020$
 & $0.000$ & $0.005$ & $0.000$ & $0.020$ \\
Flat-TabPFN$^{\dagger}$ & $7\mathrm{M}$ & $0.000$ & $0.016$ & $0.039$ & $0.106$
 & $0.000$ & $0.016$ & $0.039$ & $0.106$ \\
\midrule
Qwen3 & $8\mathrm{B}$  & $0.000$ & $0.043$ & $0.096$ & $0.191$ & $0.020$ & $0.140$ & $0.311$ & $0.373$ \\
Qwen3 & $14\mathrm{B}$ & $0.000$ & $0.067$ & $0.230$ & $0.266$ & $0.011$ & $0.128$ & $0.357$ & $0.398$ \\
Qwen3 & $32\mathrm{B}$ & $0.012$ & $0.127$ & $0.320$ & $0.407$ & $0.028$ & $0.187$ & $0.436$ & $0.527$ \\
\midrule
\rowcolor[HTML]{E7F3F9}
\textbf{TabWM} & \textbf{$\approx\!24.7\mathrm{M}$}
 & \textbf{$\mathbf{0.043}$} & \textbf{$\mathbf{0.302}$} & \textbf{$\mathbf{0.524}$} & \textbf{$\mathbf{0.687}$}
 & \textbf{$\mathbf{0.029}$} & \textbf{$\mathbf{0.302}$} & \textbf{$\mathbf{0.518}$} & \textbf{$\mathbf{0.678}$} \\
\bottomrule[1.0pt]
\end{tabular}
\end{adjustbox}
\end{center}
\vspace{-10pt}
\end{table}

\section{Identifying the Active Dynamics}
\label{sec:active_dynamics}

\begin{wrapfigure}{R}{0.37\textwidth}
    \centering
    \vspace{-3pt}
    \includegraphics[width=\linewidth]{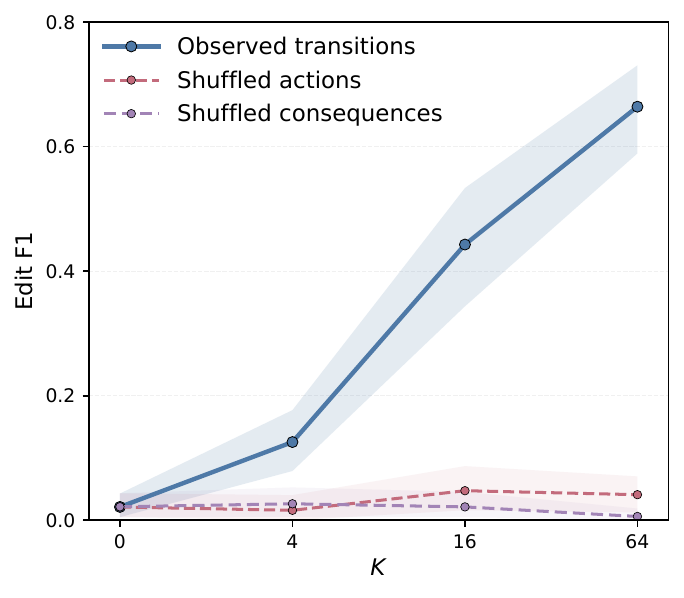}
    \vspace{-10pt}
    \caption{Behavioral context helps only when action-consequence correspondence is preserved.}
    \label{fig:execution_correspondence}
    \vspace{-10pt}
\end{wrapfigure}

Improvement from demonstrations alone does not establish that a model has inferred how actions behave in the current world. In-context learners can benefit from demonstrations even when their labels are randomized~\citep{min2022rethinking}. If TabWM is using executions as behavioral evidence, however, preserving which action produced which consequence should be essential.

We test this using the paired worlds from \S\ref{sec:dott}, where an identical observed state and action can produce different consequences under different hidden dynamics. TabWM receives either intact transitions or corrupted versions of the same observed executions. Shuffling the actions preserves the states and consequences but assigns each consequence to a different action. Shuffling the consequences preserves the states and actions but attaches the wrong response to each execution. Both interventions retain the observed components while breaking their correspondence.

With $64$ observed executions, intact transitions yield $0.664$ Edit F1, compared with $0.041$ after shuffling actions and $0.006$ after shuffling consequences (Figure~\ref{fig:execution_correspondence}). The context gain therefore depends on observing which consequences followed which actions, rather than simply on seeing additional database states or edit sets.

Which executions are most informative for a query? At fixed $K=16$, Edit F1 rises from $0.048$ when the queried action is absent from context to $0.403$ with one execution of that action and $0.700$ with four. This gain is not explained by replaying previous consequences. An explicit per-action template-transfer baseline reaches only $0.169$ and $0.216$ with one and four matching executions. TabWM therefore uses repeated executions of the queried action to infer how its effect transfers to a new relational state (\S~\ref{sec:action_coverage}).

We next vary which part of the world changes relative to the pretraining setting. A \emph{held-out state} changes the database state while leaving the surrounding world construction unchanged. A \emph{permuted action map} retains the action vocabulary but reassigns which transition mechanism each action invokes. An \emph{unseen schema} changes the relational structure, while a \emph{held-out mechanism composition} changes which familiar mechanisms occur together in one world.

\begin{wraptable}{r}{0.35\textwidth}
    \vspace{-15pt}
    \footnotesize
    \setlength{\tabcolsep}{3pt}
    \caption{Edit F1 under changes to the evaluated world.}
    \label{tab:world_changes}
    \begin{center}
    \begin{adjustbox}{max width=\linewidth}
    \begin{tabular}{@{}l|cc@{}}
        \toprule[0.8pt]
         & {\boldmath $K=16$} & {\boldmath $K=64$} \\
        \midrule
        Held-out state       & $0.483$ & $0.644$ \\
        Permuted action map  & $0.378$ & $0.604$ \\
        Unseen schema        & $0.485$ & $0.687$ \\
        Held-out composition & $0.465$ & $0.698$ \\
        \bottomrule[1.0pt]
    \end{tabular}
    \end{adjustbox}
    \end{center}
    \vspace{-15pt}
\end{wraptable}

The permuted action map is particularly diagnostic because the action identities remain unchanged while their effects do not. Edit F1 rises from $0.378$ with $16$ executions to $0.604$ with $64$, showing that behavioral context can override the mapping suggested by prior experience and recover the remapped action behavior. Additional executions also improve prediction under the other shifts (Table~\ref{tab:world_changes}).

These shifts still draw their transition mechanisms from the pretraining library. Appendix~\ref{sec:ood_dynamics} tests a stricter boundary with mechanisms introduced only at evaluation time. Behavioral context continues to improve effect localization, while complete-edit transfer depends more strongly on the quantitative transformation.

\section{Transfer to Real-Derived Structure and Values}
\label{sec:real_transfer}

The experiments above vary the active dynamics. We next vary the relational observations through which those dynamics must be inferred. Relational topology and value distributions are derived from real databases while the action and transition mechanism remain fixed.

\begin{wraptable}{r}{0.55\textwidth}
    \vspace{-10pt}
    \footnotesize
    \caption{Prediction under real-derived value distributions and relational topology.}
    \label{tab:real_transfer}
    \begin{center}
    \setlength{\tabcolsep}{6pt}
    \begin{adjustbox}{max width=\linewidth}
    \begin{tabular}{cc|cc|cc}
        \toprule[0.8pt]
        \multirow{2}{*}{\textbf{Topology}} &
        \multirow{2}{*}{\textbf{Values}} &
        \multicolumn{2}{c|}{\textbf{Edit F1}} &
        \multicolumn{2}{c}{\textbf{Support F1}} \\
        \cmidrule(lr){3-4}\cmidrule(lr){5-6}
        & & $K=16$ & $K=64$ & $K=16$ & $K=64$ \\
        \midrule
        Synthetic    & Synthetic    & $0.481$ & $0.696$ & $0.675$ & $0.875$ \\
        Synthetic    & Real-derived & $0.355$ & $0.522$ & $0.614$ & $0.789$ \\
        Real-derived & Real-derived & $0.365$ & $0.497$ & $0.661$ & $0.767$ \\
        \bottomrule[1.0pt]
    \end{tabular}
    \end{adjustbox}
    \end{center}
    \vspace{-12pt}
\end{wraptable}

We first replace the relational structure. Table and primary- and foreign-key topology are imported from $139$ databases in THE JOIN~\citep{ranjan2026largescale} and paired with $139$ size-matched synthetic references under the same protocol. With $64$ observed executions, Edit F1 is $0.657$ on the imported topologies and $0.671$ on the matched synthetic worlds. Replacing relational connectivity alone therefore causes little loss.

We then change the values populating those structures. We derive empirical value distributions from $241$ columns across six RelBench databases~\citep{robinson2024relbench} and map their marginal shapes into the ranges supported by the generator. A separate $200$-world evaluation compares synthetic and real-derived values, first under synthetic topology and then with real-derived topology as well.

The larger transfer gap appears when the value distributions change (Table~\ref{tab:real_transfer}). Edit F1 decreases under real-derived values, and Support F1 falls as well, showing that the shift affects where consequences are predicted to occur rather than only the reconstruction of numeric values. More behavioral context remains beneficial in every condition. Adding real-derived topology to the value shift produces little further degradation. Across these observation shifts, value distributions therefore matter more than relational connectivity.

\section{From Predicted Consequences to Decisions}
\label{sec:decisions}

Comparing actions requires more than predicting one consequence at a time. A useful world model must imagine possible futures and distinguish which are preferable before acting. We therefore use TabWM as a rollout model: predicted edits are applied recursively along candidate action sequences, and the resulting trajectories are scored with the known task objective~\citep{hafner2019planet,chua2018pets}. We first ask whether these imagined futures preserve the ordering induced by the true system, and then whether that ordering leads to better choices when more alternatives are available.

\subsection{Ranking Candidate Futures}
\label{sec:future_ranking}

At each decision state, we sample a shared pool of $64$ legal three-step action sequences. TabWM rolls each sequence forward and scores the imagined trajectory using the known world-specific objective, which combines changes in state potential with action costs (\S\ref{sec:planning_protocol}). We execute the same sequences in the true simulator to obtain their realized returns. The evaluation contains $160$ decision states from $40$ held-out worlds, with states and candidate pools shared across context sizes. Spearman correlation is computed within each $64$-sequence pool and averaged across decision states. Normalized regret measures, within the same pool, the realized-return gap between the selected candidate and the best available candidate, divided by the pool's return span.

To separate information about state changes from simpler sources of ranking signal, we use three references. The \emph{simulator oracle} rolls candidates forward with the true transition mechanism rather than TabWM. The \emph{shuffled-context} control mismatches demonstrated actions with their observed consequences before prediction, preserving the amount of context while breaking its behavioral correspondence. The \emph{no-change} control keeps the predicted state fixed, leaving action costs as the only source of variation between sequences.

\begin{wraptable}{r}{0.46\textwidth}
\vspace{-10pt}
\footnotesize
\caption{Candidate ranking under TabWM and reference rollouts.}
\label{tab:future_ranking}
\begin{center}
\begin{adjustbox}{max width=\linewidth}
\begin{tabular}{lc|cc}
\toprule[0.8pt]
\textbf{Scorer} & {\boldmath $K$} & \textbf{Spearman $\rho$} & \textbf{Normalized regret} \\
\midrule
Simulator oracle & -- & $0.732$ & $0.107$ \\
No-change        & -- & $0.345$ & $0.393$ \\
Shuffled context & $64$ & $0.338$ & $0.392$ \\
\midrule
\multirow{3}{*}{\textbf{TabWM}}
& $4$  & $0.444$ & $0.298$ \\
& $16$ & $0.590$ & $0.208$ \\
& $64$ & $0.622$ & $0.207$ \\
\bottomrule[1.0pt]
\end{tabular}
\end{adjustbox}
\end{center}
\vspace{-10pt}
\end{wraptable}

With intact behavioral context, TabWM's imagined ranking becomes increasingly consistent with realized returns (Table~\ref{tab:future_ranking}). With $64$ observed executions, its Spearman correlation reaches $0.622$, compared with $0.338$ after shuffling the action--consequence correspondence and $0.345$ when state changes are ignored altogether. The simulator oracle reaches $0.732$. The contrast with the two controls shows that the ranking gain comes from information about how candidate actions change the current world, rather than only from their immediate costs.

The full ranking continues to improve beyond $16$ executions, but most of the gain in the selected candidate has already appeared. Normalized regret falls from $0.298$ with four executions to $0.208$ with $16$, and changes little thereafter. One-step consequence prediction, by contrast, continues to improve at larger context sizes. In this planning task, the behavioral evidence needed to distinguish good candidates therefore appears to saturate earlier than the evidence needed to reconstruct their consequences in detail. This separation echoes prior observations that predictive fidelity and control utility need not improve in lockstep~\citep{lambert2020mismatch,grimm2020value,zhang2026active}.

\subsection{Choosing Among More Futures}
\label{sec:candidate_search}

\begin{wraptable}{r}{0.38\textwidth}
\renewcommand{\arraystretch}{1.25}
\footnotesize
\caption{Median realized return of selected sequences as the candidate budget grows at $K=16$.}
\label{tab:candidate_search}
\begin{center}
\begin{adjustbox}{max width=\linewidth}
\begin{tabular}{l|cccc}
\toprule[0.8pt]
 & {\boldmath $N=1$} & {\boldmath $N=4$} & {\boldmath $N=16$} & {\boldmath $N=64$} \\
\midrule
Simulator oracle   & $-1.019$ & $-0.348$ & $-0.078$ & $0.285$ \\
\midrule
\textbf{TabWM}     & $-1.019$ & $-0.462$ & $-0.323$ & $-0.165$ \\
Shuffled context   & $-1.019$ & $-0.668$ & $-0.513$ & $-0.412$ \\
No-change          & $-1.019$ & $-0.646$ & $-0.508$ & $-0.398$ \\
\bottomrule[1.0pt]
\end{tabular}
\end{adjustbox}
\end{center}
\vspace{-15pt}
\end{wraptable}

Once candidate futures can be ranked meaningfully, search determines how much that ranking can be exploited. We fix the behavioral context and reveal progressively larger nested subsets of the same $64$ legal sequences, so broader search adds choices without changing those already available.

The common starting point in Table~\ref{tab:candidate_search} is a useful sanity check: before search creates a choice, every scorer must agree. As the candidate pool grows, the rankings begin to matter. TabWM consistently selects higher-return sequences than either control, while the simulator oracle shows the remaining value of more accurate dynamics. The controls also improve because action costs alone induce a weak preference over candidates.

This advantage is not driven only by a few large gains. Across the final expansion of the candidate pool, TabWM improves the selected return in $30$ of $40$ held-out worlds, compared with $21$ under shuffled context and $22$ under no-change dynamics. Taken together, the two planning experiments show that behavioral context and search play complementary roles. Executions provide an informative ordering over candidate futures, while broader search gives that ordering more opportunities to improve the decision.

\section{Related Work}
\label{sec:related_work}

\textbf{Relational foundation models.}
Relational learning has recently moved from task-specific models toward reusable models that transfer across databases and prediction tasks. Griffin~\citep{wang2025griffin}, the Relational Transformer~\citep{ranjan2026relational}, and KumoRFM-2~\citep{hudovernik2026kumorfm2} learn transferable representations directly over relational data, while synthetic pretraining and in-context formulations further support adaptation across databases~\citep{kothapalli2026plurel,wang2026relational,hayler2026relarena}. Despite these different approaches, the learning interface remains largely predictive, with relational context used to infer targets or properties of database records. TabWM changes what is inferred from context: observed executions provide evidence about how actions change those records.

\textbf{Learning dynamics from behavioral context.}
Once the target becomes an action-conditioned transition, our setting connects to work on learning action effects and identifying dynamics from experience. Relational action models learn structured or stochastic effects for planning~\citep{pasula2007symbolic,martinez2016relational}, while context-conditioned dynamics models adapt prediction from recent transitions~\citep{lee2020context}. Prior-data fitted learning provides a broader route to amortizing such inference across tasks~\citep{muller2022pfn}, including models trained across synthetic dynamical systems that infer new behavior from input-output histories~\citep{forgione2023classmodels,rufolo2024enhanced}. TabWM brings this form of identification to relational systems, where an action can induce sparse changes that propagate across linked records.

\textbf{World models for operational systems.}
In operational systems, the transition mechanism itself may not be apparent from the values stored in the database~\citep{klein2025position,klein2026statistically}. Enterprise workflow benchmarks expose cascading action effects~\citep{gupta2026workflows}, while process simulators recover executable models from event logs for what-if analysis~\citep{chapelacampa2025simod}. When the active configuration is readable, runtime discovery provides another way to ground predictions in the current system~\citep{nair2026enterprise}. TabWM studies the complementary case in which the mechanism is hidden but its executions are observable, allowing behavior in the current world to be inferred from context.

\section{Conclusion}

Relational databases can describe not only what a system looks like, but also how it changes. TabWM learns from observed executions to predict sparse relational consequences of new actions, adapting through context rather than parameter updates. Because these predictions are explicit edits, they can be composed into candidate futures and used to compare actions before they are taken. More broadly, our results suggest that relational dynamics can be learned in context. Pretraining across executable worlds allows a single model to infer action consequences from executions in a new world.

\paragraph{Limitations \& Future Work.}
Moving from relational prediction to relational world modeling also changes what is required from an evaluation benchmark. Studying behavior from experience requires aligned observations of the relational state before an action, the action itself, and the resulting changes, which current relational benchmarks rarely expose. This motivates our use of executable worlds and suggests a natural next step in building datasets around real operational execution traces. Such benchmarks would enable broader evaluation across systems and open the study of how relational models infer, track, and use dynamics from experience.

\begin{ack}
\end{ack}

\bibliographystyle{plainnat}
\bibliography{reference}


\newpage
\appendix
\input{supplementary_results.tex}

\end{document}

%% file: supplementary_results.tex

\definecolor{appblue}{HTML}{4C78A8}
\definecolor{apporange}{HTML}{E89C4A}
\definecolor{appgreen}{HTML}{59A14F}
\definecolor{apppurple}{HTML}{8E6C8A}
\definecolor{appgray}{HTML}{6F7782}

\tikzset{
    statebox/.style={
        draw=appblue,
        fill=appblue!7,
        rounded corners=2pt,
        line width=.55pt,
        inner sep=5pt,
        align=left
    },
    actionbox/.style={
        draw=apporange,
        fill=apporange!9,
        rounded corners=5pt,
        line width=.55pt,
        inner sep=5pt,
        align=center
    },
    effectbox/.style={
        draw=appgreen,
        fill=appgreen!8,
        rounded corners=2pt,
        line width=.55pt,
        inner sep=5pt,
        align=left
    },
    latentbox/.style={
        draw=apppurple,
        fill=apppurple!8,
        rounded corners=2pt,
        line width=.55pt,
        inner sep=5pt,
        align=center
    },
    apparrow/.style={
        -{Latex[length=1.8mm]},
        draw=black!55,
        line width=.6pt
    },
    relarrow/.style={
        -{Latex[length=1.8mm]},
        draw=appblue!75,
        line width=.6pt
    }
}

\newtcolorbox{promptbox}{
    enhanced,
    sharp corners,
    colback=black!2,
    colframe=black!30,
    boxrule=0.45pt,
    left=2mm,
    right=2mm,
    top=1.5mm,
    bottom=1.5mm
}

\section{Extended Discussion}
\label{sec:extended_discussion}

Relational foundation models have largely expanded pretraining along the diversity of databases, schemas, and prediction tasks~\citep{ranjan2026relational,kothapalli2026plurel,wang2026relational}. World modeling introduces another axis: the diversity of behaviors that relate actions to future relational states. Scaling along this axis is not simply a matter of observing more database instances. It requires a prior that captures reusable patterns of change while allowing their concrete realization to vary across systems. This suggests behavioral diversity as a distinct dimension of relational pretraining alongside structural and semantic diversity.

A broad prior describes which behaviors are plausible, but evidence from the current system is still needed to determine which are active. Observed executions provide one such source by directly connecting actions to realized consequences~\citep{forgione2023classmodels,rufolo2024enhanced}. Runtime-readable configuration can provide another by exposing rules instantiated in the current system~\citep{nair2026enterprise}. These signals constrain the dynamics in different ways. A pretrained prior captures regularities across systems, configuration describes explicitly available rules, and executions reveal their realized behavior under the current state. Combining these sources raises a broader question of how a relational world model should update its view of the current system when evidence differs in specificity or completeness, and how such evidence should be exposed by operational environments in the first place.

Once a model can act, the evidence available for identification is no longer fixed. Actions can change both the system and what can be learned about it, connecting planning with ideas from dual control and active identification~\citep{mesbah2018stochastic}. The relevant uncertainty, however, need not be the uncertainty that contributes most to average transition error. Our planning results show that consequence prediction and action selection improve at different rates, consistent with prior observations that predictive fidelity and control utility need not align perfectly~\citep{lambert2020mismatch,grimm2020value}. This suggests a decision-directed view of identification in which a model seeks evidence that can change which futures should be preferred.

\section{Experimental Details}
\label{sec:evaluation_details}

\subsection{Metrics}

An action consequence is represented as a set of typed edits. For an edit $e$, let
\[
\operatorname{supp}(e)
=
(o_e,t_e,r_e,c_e)
\]
denote its structural support, consisting of the operation, table, row, and column whenever the corresponding fields apply. This representation separates identifying \emph{what changes} from predicting the value produced by that change.

\paragraph{Support F1.}
Two edits match in support when their operations and relational targets agree. Support F1 therefore measures whether the model locates the consequence of an action, independently of the value written at each location. This distinction is useful in relational systems because an action may affect several records reached through different relational paths.

\paragraph{Edit F1.}
Complete-edit matching additionally requires the new value to be correct. Categorical values match exactly. Numeric values use
\begin{equation}
|\hat v-v|
\leq
\max\left(10^{-3},\,0.02|v|\right).
\label{eq:app_numeric_match}
\end{equation}
Edit F1 is computed from the matched and unmatched edits after duplicate predictions at the same structural support are collapsed.

\paragraph{Value MAE.}
For numeric updates whose structural support is matched, we additionally report
\begin{equation}
\operatorname{MAE}_{\mathrm{value}}
=
\frac{1}{|\mathcal M_{\mathrm{num}}|}
\sum_{(e,\hat e)\in\mathcal M_{\mathrm{num}}}
|v_e-\hat v_e|,
\label{eq:app_value_mae}
\end{equation}
where $\mathcal M_{\mathrm{num}}$ is the set of correctly located numeric edits. Value MAE is therefore conditional on localization and is used as a diagnostic of numeric reconstruction rather than as the primary cross-model metric.

\paragraph{Confidence intervals.}
Queries drawn from the same world share its transition dynamics. Bootstrap estimates therefore resample worlds and retain the queries associated with each sampled world. The main context sweep uses $2{,}000$ world-bootstrap draws.

\subsection{Baselines}
\label{sec:baseline_details}

All baselines are evaluated on the same sampled tasks and behavioral context as TabWM. They differ in how the relational execution is represented and in the model used to predict its consequence.

\paragraph{Flat baselines.}
Each candidate query cell becomes one row in a fixed-width feature table. Its features describe the current value and type, column-level statistics, whether the row is bound to or linked from the queried action, and summaries of demonstrated changes to the same column. Flat-GBM applies gradient-boosted trees~\citep{friedman2001greedy} to this representation, while Flat-TabPFN uses TabPFN~\citep{hollmann2025accurate}. The representation retains local values and coarse relational evidence but does not preserve each execution as a structured relational object. Table and column names are excluded, so the flat baselines are unchanged by the naming intervention.

\paragraph{LLM baselines.}
Qwen3~\citep{yang2025qwen3} receives a textual serialization of the same observed executions and query under the same action-centered cell budget as TabWM. The readable and opaque conditions differ only in the identifiers assigned to schema elements, categorical values, and actions. Relational links, values, row identities, and action bindings remain unchanged.

\subsection{LLM Prompt Construction}
\label{sec:llm_prompt}

LLMs have demonstrated strong adaptability across diverse tasks~\citep{zhang2026consolidating}, and recent work has shown that this capability extends to structured data~\citep{zhang2025grade,yang-etal-2026-sage,yang-etal-2025-doubling}. Following the relational serialization used for LLM baselines in prior relational foundation-model evaluations~\citep{ranjan2026relational}, the prompt states the schema and action signatures before presenting the observed executions. Each demonstration contains a pre-action relational neighbourhood, the executed action, and the resulting edit set. The final query uses the same format with its edits withheld.

Figure~\ref{fig:qwen_prompt} shows an abridged prompt in the readable naming condition.

\input{figures/prompt}

\subsection{Candidate-Future Scoring}
\label{sec:planning_protocol}

Each world defines a state potential $\Phi(S)$ and an action cost $c(A)$. For a horizon-$H$ candidate sequence, the predicted return is
\begin{equation}
\widehat R(A_{0:H-1})
=
\sum_{h=0}^{H-1}
\gamma^h
\left[
\Phi(\widehat S_{h+1})
-
\Phi(\widehat S_h)
-
c(A_h)
\right],
\qquad
H=3,\ \gamma=0.95.
\label{eq:app_plan_score}
\end{equation}
The predicted state is updated after every imagined action using Equation~\ref{eq:tabwm-rollout}, while the behavioral context remains fixed throughout the rollout.

The evaluation contains four decision states from each of $40$ held-out worlds. Decision states are reached through actions sampled independently of all scorers. At each state, we sample one shared pool of $64$ legal action sequences, which is reused across context sizes and scoring methods. The candidate-budget experiment uses nested prefixes of this same pool for $N\in\{1,4,16,64\}$.

The simulator oracle replaces TabWM with the true transition mechanism. Shuffled context breaks the action--consequence correspondence in the observed executions, while the no-change scorer leaves the predicted state fixed and therefore uses only action costs. Spearman correlation is computed within each $64$-sequence pool and averaged across the $160$ decision states. Normalized regret divides the selected candidate's true-return shortfall from the best candidate by the true-return span of that pool.

\section{Method Details}
\label{sec:model_details}

TabWM must preserve two kinds of structure at once: the relational structure of the database and the correspondence between each observed action and its consequence~\citep{ranjan2026relational}. The encoder represents both in a shared context, while the decoder predicts the query consequence as a set of typed edits and transfers numeric transformations across executions.

\subsection{Relational Transition Encoding}
\label{sec:transition_encoding_details}

The relational state is represented with the pretrained Relational Transformer backbone~\citep{ranjan2026relational}, using weights learned from \textsc{PluRel}-generated relational databases~\citep{kothapalli2026plurel}. TabWM extends this representation from a static relational neighborhood to an execution by placing the state, action, and observed consequence in the same relational context.

For each execution, an action-centered neighborhood is collected by following primary- and foreign-key links from the records referenced by the action until the cell budget is reached. Cells retain their column identity, type, and value. Textual identifiers use deterministic hash embeddings rather than a pretrained language encoder. This lets the model preserve identity within a task without introducing external linguistic semantics.

The action is represented as a distinguished pseudo-row linked to the records bound by its entity arguments. Scalar and categorical arguments are represented as action cells. Demonstrated edits are attached to the records they modify. Role embeddings distinguish state, action, and consequence cells, while transition identifiers mark which elements belong to the same observed execution. For an updated cell, shared table, row, and column identities align its before-action representation with the edit that follows the action.

These connections preserve both relational structure and execution membership. When the demonstrations and query are packed as in Equation~\ref{eq:transition-encoding}, the query consequence is withheld while its state and action share attention with the observed executions. The encoder can therefore relate a queried action to the current relational state and to before-and-after evidence from the same world.

\subsection{Structured Edit Prediction}
\label{sec:structured_edit_prediction_details}

An action may change several relational targets or none, and the order of those changes has no meaning. TabWM therefore predicts the consequence as an unordered set using learned edit queries, following DETR-style set prediction~\citep{carion2020end}. Each query attends to the contextual representation and predicts whether an edit exists, which operation it performs, where it acts, and any value or record contents required by that operation.

The predicted operation determines the form of the relational target. Value and state updates point to affected cells, while row- and table-level operations select the corresponding relational objects. Typed output heads predict categorical and numeric contents. The decoder therefore separates identifying \emph{what changes} from predicting \emph{what value results}, while keeping both parts within the same edit representation.

There is no canonical ordering between predicted slots and reference edits. During training, minimum-cost bipartite matching assigns predicted slots to reference edits using their operation, relational target, and value when applicable. Matched slots receive supervision for the corresponding edit fields, while unmatched slots learn to emit no edit. The supervision is applied to the observed consequence itself. TabWM is not given the hidden transition program that produced it.

\subsection{Numeric Value Transfer}
\label{sec:numeric_decoder_details}

Once a numeric target has been identified, the remaining question is how its value should change. The decoder uses visible before-and-after pairs $(x_r,y_r)$ from previous executions that modify the same relational role. Here $x_r$ is the value before the demonstrated action and $y_r$ is the value written by its consequence. The edit query, target representation, and query action condition attention over these pairs, allowing the model to select changes that are relevant to the current query rather than treating all observed numeric edits equally.

A learned path first predicts a numeric estimate without assuming a particular transformation form. It produces an initial estimate $\hat y_0$ together with a residual correction,
\begin{equation}
\hat y_{\mathrm{learned}}
=
\hat y_0+r_{\theta}.
\label{eq:app_learned_numeric}
\end{equation}
The contextual representation used for this prediction includes the selected before-and-after evidence and the actions under which those changes were observed.

Alongside this learned estimate, TabWM constructs two transformation candidates from the same evidence. Let $\alpha_r$ denote normalized attention weights over usable pairs and let $\bar x$ and $\bar y$ be their weighted means. For a query value $x_q$,
\begin{equation}
h_{\mathrm{add}}
=
x_q+\sum_r\alpha_r(y_r-x_r),
\qquad
h_{\mathrm{aff}}
=
\bar y+\beta(x_q-\bar x),
\label{eq:app_numeric_candidates}
\end{equation}
where $\beta$ is a bounded weighted least-squares slope. The additive candidate transfers an observed change, while the affine candidate estimates how the outcome varies with the starting value. Both therefore apply a transformation inferred from the demonstrations to the query value rather than copying a demonstrated outcome.

Learned gates combine these candidates with the neural estimate as in Equation~\ref{eq:numeric-composition}. This gives the decoder several ways to use the same behavioral evidence. The learned path can represent changes that are not captured by a simple arithmetic form, while the additive and affine paths provide direct routes for regularities that transfer cleanly across starting values.

Candidate construction uses only visible before-and-after pairs that remain informative after value-domain clipping. The affine candidate additionally requires at least two usable pairs with variation in their starting values. When a candidate is unavailable, it contributes no correction to $\hat y_{\mathrm{learned}}$. If no usable numeric evidence is present, prediction falls back to the learned estimate. These choices affect only the value written by a numeric edit and do not alter its predicted operation or relational target. The arithmetic paths only modify the value attached to an already selected numeric edit. \S~\ref{sec:numeric_reader_ablation} isolates their contribution under matched reader training.

\subsection{Training Details}
\label{sec:training_details}

TabWM initializes its relational encoder from the pretrained Relational Transformer used in \S\ref{sec:transition_encoding}. The encoder, structured edit decoder, and numeric value decoder are then optimized jointly on tasks sampled from \textsc{WorldRel}. Each task contains a behavioral context and a separate query from the same world, with the context size sampled from $K\in\{0,1,2,4,8,16,32,64\}$. Training, development, and evaluation use disjoint world seed ranges.

For numeric edits, training uses a routing curriculum to provide a stable target representation while the edit pointers are still inaccurate. After set matching, a matched numeric slot may initially read the reference target cell when constructing the value-decoder input. This probability is annealed to zero during training, after which the model uses its own predicted target throughout. Matching is computed before the route is chosen, and validation, evaluation, and rollout always use predicted targets.

\begin{table}[!ht]
\footnotesize
\caption{Training details for TabWM.}
\label{tab:training_details}
\begin{center}
\begin{tabular}{l|l}
\toprule
Initialization & Relational Transformer pretrained on PluRel~\citep{kothapalli2026plurel} \\
Trainable components & Relational encoder, edit decoder, numeric value decoder \\
Context sizes & $K\in\{0,1,2,4,8,16,32,64\}$ \\
Optimizer & AdamW \\
Learning rates & $7.5{\times}10^{-5}$ encoder; $3{\times}10^{-4}$ transition components \\
Training length & 200k optimizer updates \\
Routing curriculum & Reference-to-predicted target over the first 50k updates \\
Checkpoint selection & Highest pooled development Edit F1 at $K\in\{16,64\}$ \\
\bottomrule
\end{tabular}
\end{center}
\end{table}

\section{A Formal View of Behavioral Identification}
\label{sec:behavioral_identification_view}

Observed executions can be interpreted as evidence about a latent transition rule $T_w$. This connects TabWM to in-context system identification~\citep{forgione2023classmodels,rufolo2024enhanced} and to prior-data fitted learning, where inference over a task-specific latent structure is amortized across a pretraining distribution~\citep{muller2022pfn}. This section makes that interpretation explicit without assuming that TabWM recovers a symbolic transition program.

\subsection{Posterior-Predictive Interpretation}

The behavioral context
\[
\mathcal C_w^K
=
\{(S_i,A_i,\Delta S_i)\}_{i=1}^{K}
\]
provides evidence about $T_w$. A posterior-predictive interpretation of Equation~\ref{eq:behavioral-inference} is
\begin{equation}
p(\Delta S_q\mid S_q,A_q,\mathcal C_w^K)
=
\mathbb E_{T\sim p(T\mid\mathcal C_w^K)}
\left[
p(\Delta S_q\mid S_q,A_q,T)
\right].
\label{eq:posterior_predictive_dynamics}
\end{equation}
TabWM does not explicitly recover or parameterize the posterior over symbolic transition programs. Instead, pretraining across worlds learns the resulting predictive map from behavioral context to query consequences.

Under this view, readable identifiers may affect the prior over plausible dynamics, while observed executions constrain that prior through realized action--consequence pairs. The naming intervention in \S\ref{sec:identification} changes the former while preserving the latter.

\subsection{Identification Through Correspondence}

For intuition, suppose the dynamics are deterministic and define
\begin{equation}
\mathcal T(\mathcal C_w^K)
=
\left\{
T:
T(S_i,A_i)=\Delta S_i
\quad
\forall(S_i,A_i,\Delta S_i)\in\mathcal C_w^K
\right\}
\label{eq:compatible_dynamics}
\end{equation}
as the set of transition rules compatible with the observed executions. A query is identified whenever all compatible rules agree on its consequence,
\begin{equation}
\left|
\left\{
T(S_q,A_q):
T\in\mathcal T(\mathcal C_w^K)
\right\}
\right|
=1.
\label{eq:query_identified}
\end{equation}

Each aligned action--consequence pair rules out transition functions that are inconsistent with the observed execution. Breaking this correspondence removes those constraints because the resulting pairs need not be jointly explained by a single $T$. The shuffling intervention in \S\ref{sec:active_dynamics} tests this consequence of the identification view directly.

\section{The WorldRel Prior}
\label{sec:worldrel_prior}

A relational world is defined not only by what its database looks like, but by how that database changes under actions. \textsc{WorldRel} turns this distinction into a generative prior. Starting from relational databases sampled with \textsc{PluRel}~\citep{kothapalli2026plurel}, it adds executable actions and world-specific transition mechanisms whose effects remain consistent within a world and vary across worlds. The resulting tasks expose TabWM to different relational structures and different ways in which those structures can evolve.

\subsection{Generating Executable Worlds}
\label{sec:worldrel_generation}

Each sampled world fixes four components: a relational schema, an action interface, an initial-state distribution, and a transition mechanism. The base relational schema contains between $2$ and $12$ sampled tables with typed attributes and sampled foreign-key structure. Each world also contains between $4$ and $24$ actions that bind a primary entity and may additionally take entity, scalar, or categorical arguments. Actions can carry state-dependent preconditions and costs. Resetting the world draws a fresh relational state while preserving its action interface and dynamics.

\begin{table}[!ht]
\centering
\footnotesize
\caption{Sources of variation in a \textsc{WorldRel} world.}
\label{tab:worldrel_components}
\begin{center}
\begin{adjustbox}{max width=.9\linewidth}
\begin{tabular}{l|l}
\toprule[0.8pt]
\textbf{Component} & \textbf{Variation within the prior} \\
\midrule
Relational structure
& $2$--$12$ tables, typed columns, and sampled foreign-key topology \\

Initial state
& Fresh relational states drawn when an episode is reset \\

Action interface
& $4$--$24$ actions with entity, scalar, and categorical arguments \\

Transition mechanism
& Ordered compositions drawn from $13$ executable effect classes \\
\bottomrule[1.0pt]
\end{tabular}
\end{adjustbox}
\end{center}
\vspace{-10pt}
\end{table}

Actions are governed by compositions of executable effects. The transition library spans local numeric updates and state transitions, propagation through foreign-key links, multi-row cascades, aggregate recomputation, threshold-triggered changes, stochastic and regime-dependent behavior, event creation, deletion, and delayed consequences. Relational effects can follow foreign-key paths for up to three hops and may fan out across several linked records. Delayed effects can arrive one to four steps after the action that scheduled them.

Several effects may occur within the same action. They are applied in sequence to the evolving state, allowing one change to alter the conditions or inputs of another. Their target columns, coefficients, thresholds, transition maps, delays, and stochastic parameters remain fixed within a world and are resampled across worlds. Direct numeric updates include additive, multiplicative, reset, and action-conditioned cross-table transformations, while aggregate effects derive values from linked records through \texttt{sum}, \texttt{mean}, \texttt{max}, or \texttt{count}. A consequence can therefore combine value changes with relational propagation, state transitions, event creation, and structural effects.

A realized sample of $1{,}500$ worlds from the training seed range contains $14.0$ actions and $2.17$ effects per action on average, producing $1.38$ edits per transition. Importantly, $38.1\%$ of realized edits occur on a table other than the one directly bound by the action. A separate $500$-world sample from the evaluation range gives nearly identical statistics. The executable prior therefore produces substantial relational propagation in practice rather than only through rare constructed cases.

\subsection{Variation Across Worlds and Tasks}
\label{sec:worldrel_generalization}

Sampling a new world changes several parts of the system at once. Its relational schema and initial-state distribution are resampled together with the action interface, the assignment from actions to mechanisms, and the parameters and composition of those mechanisms. The mechanism families and structured edit vocabulary provide a common behavioral prior across worlds.

\begin{table}[!ht]
\vspace{-10pt}
\footnotesize
\caption{Variation between pretraining worlds and held-out \textsc{WorldRel} worlds.}
\label{tab:worldrel_generalization}
\setlength{\tabcolsep}{7pt}
\begin{center}
\begin{adjustbox}{max width=.86\linewidth}
\begin{tabular}{l|c}
\toprule[0.8pt]
\textbf{World component} & \textbf{Held-out world} \\
\midrule
Schema and foreign-key graph
& Resampled \\

Initial-state distribution
& Resampled \\

Table, column, category, and action identities
& Resampled \\

Action argument signatures
& Resampled \\

Action-to-mechanism assignment
& Resampled \\

Mechanism parameters
& Resampled \\

Mechanism composition
& Resampled \\

Mechanism families
& Shared \\

Edit vocabulary
& Shared \\
\bottomrule[1.0pt]
\end{tabular}
\end{adjustbox}
\end{center}
\vspace{-10pt}
\end{table}

Observed executions distinguish among the behaviors made possible by this prior. An action can therefore invoke different effects in different worlds, while related mechanisms can appear under different schemas, parameters, and relational bindings. The shifts in \S\ref{sec:active_dynamics} probe these sources of variation separately by changing the action mapping, schema, or mechanism composition. Throughout the main evaluation, an unseen world is a fresh instantiation of this executable-world prior.

A task draws its observed executions and query from independently reset episodes of the same world. Each world provides $20$ episodes of four steps, giving up to $80$ transitions from the same underlying dynamics. During pretraining, the amount of behavioral context is sampled from $K\in\{0,1,2,4,8,16,32,64\}$. Drawing context across episodes exposes repeated behavior from the same world without giving the query its own execution history.

World generation is also separated from rendering. Schema, categorical, and action identifiers can be replaced consistently with opaque labels while preserving the relational structure, values, row identities, action bindings, and consequences. The naming experiments in \S\ref{sec:identification} use this separation to vary semantic cues without changing the underlying behavioral task.

\subsection{Behavior Transfer Across Executions}
\label{sec:execution_example}

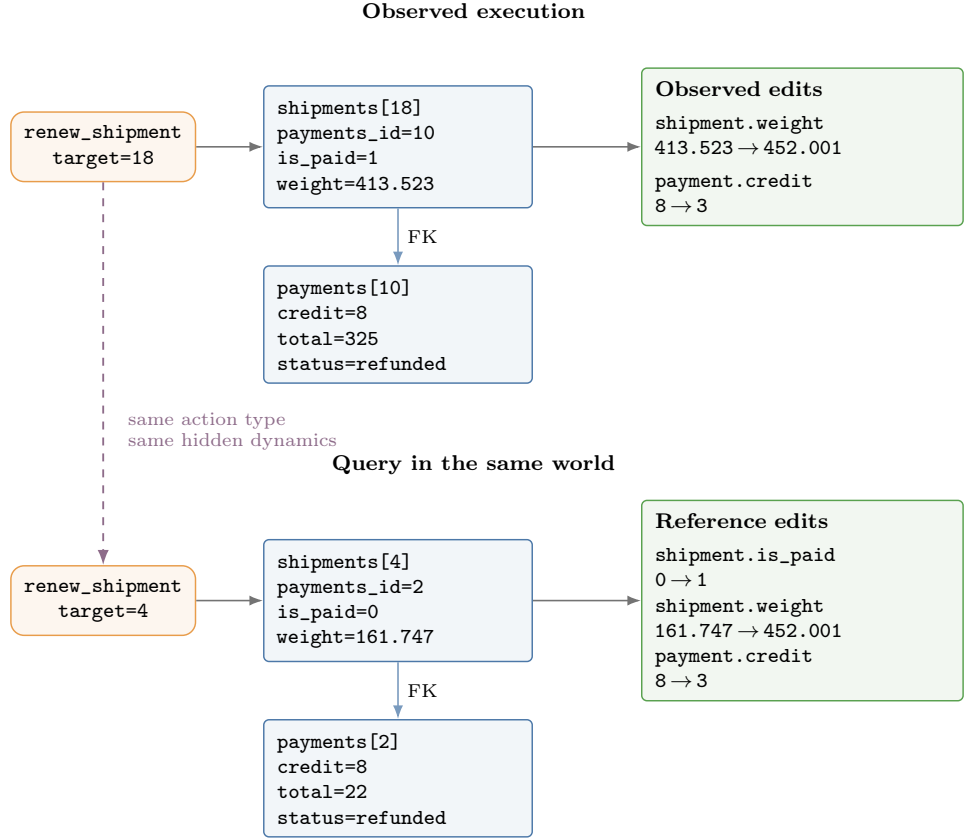
\begin{figure}[!htbp]
\centering
\begin{adjustbox}{max width=.8\linewidth}
\begin{tikzpicture}[font=\footnotesize]

\node[font=\footnotesize\bfseries] at (6.4,4.1) {Observed execution};

\node[actionbox] (a1) at (1.5,2.3)
{\texttt{renew\_shipment}\\
\texttt{target=18}};

\node[statebox, text width=32mm] (s1) at (5.4,2.3)
{\textbf{\texttt{shipments[18]}}\\
\texttt{payments\_id=10}\\
\texttt{is\_paid=1}\\
\texttt{weight=413.523}};

\node[statebox, text width=32mm] (p1) at (5.4,-0.05)
{\textbf{\texttt{payments[10]}}\\
\texttt{credit=8}\\
\texttt{total=325}\\
\texttt{status=refunded}};

\node[effectbox, text width=39mm] (e1) at (10.75,2.3)
{\textbf{Observed edits}\\[1mm]
\texttt{shipment.weight}\\
\texttt{413.523}\,$\rightarrow$\,\texttt{452.001}\\[1mm]
\texttt{payment.credit}\\
\texttt{8}\,$\rightarrow$\,\texttt{3}};

\draw[apparrow] (a1.east) -- (s1.west);
\draw[relarrow] (s1.south) -- node[right,font=\scriptsize]{FK} (p1.north);
\draw[apparrow] (s1.east) -- (e1.west);

\node[font=\footnotesize\bfseries] at (6.4,-1.9) {Query in the same world};

\node[actionbox] (a2) at (1.5,-3.7)
{\texttt{renew\_shipment}\\
\texttt{target=4}};

\node[statebox, text width=32mm] (s2) at (5.4,-3.7)
{\textbf{\texttt{shipments[4]}}\\
\texttt{payments\_id=2}\\
\texttt{is\_paid=0}\\
\texttt{weight=161.747}};

\node[statebox, text width=32mm] (p2) at (5.4,-6.05)
{\textbf{\texttt{payments[2]}}\\
\texttt{credit=8}\\
\texttt{total=22}\\
\texttt{status=refunded}};

\node[effectbox, text width=39mm] (e2) at (10.75,-3.7)
{\textbf{Reference edits}\\[1mm]
\texttt{shipment.is\_paid}\\
\texttt{0}\,$\rightarrow$\,\texttt{1}\\
\texttt{shipment.weight}\\
\texttt{161.747}\,$\rightarrow$\,\texttt{452.001}\\
\texttt{payment.credit}\\
\texttt{8}\,$\rightarrow$\,\texttt{3}};

\draw[apparrow] (a2.east) -- (s2.west);
\draw[relarrow] (s2.south) -- node[right,font=\scriptsize]{FK} (p2.north);
\draw[apparrow] (s2.east) -- (e2.west);

\draw[-{Latex[length=2mm]}, dashed, line width=.7pt, apppurple]
(a1.south) -- node[pos=.65, right=2mm, font=\scriptsize, text=apppurple, align=left]
{same action type\\same hidden dynamics} (a2.north);

\end{tikzpicture}
\end{adjustbox}

\caption{Transferring behavior across executions. The demonstrated action and the query bind to different shipment and payment rows, so the relational targets must be transferred by role rather than by row identity. The query additionally changes \texttt{is\_paid} because its starting value differs from the demonstration. Replaying the demonstrated edit list would therefore be insufficient.}
\label{fig:worked_execution}
\vspace{-8pt}
\end{figure}

The same world-level behavior can produce different concrete edits when an action is applied in a different relational state. Figure~\ref{fig:worked_execution} shows one observed execution and a query from the same world. They share an action type and transition mechanism, but bind to different records and begin from different values.

The observed action changes both the targeted shipment and its linked payment. In the query, the same relational roles are occupied by different rows, so replaying the demonstrated row identities would fail. The starting state also differs, producing an additional edit to \texttt{is\_paid}. What transfers across executions is therefore not a literal edit list, but the relationship between an action, the current relational state, and the records and values it changes.

\section{Additional Analyses}
\label{sec:additional_results}

\subsection{Held-Out Confirmation}
\label{sec:confirmation_results}

A larger confirmation set of $512$ unseen worlds reproduces the main context-scaling pattern. Support and complete-edit prediction both improve as more executions become available, while conditional numeric error decreases.

\begin{table}[!ht]
\footnotesize
\caption{Held-out confirmation on $512$ unseen worlds.}
\label{tab:confirmation}
\setlength{\tabcolsep}{8pt}
\begin{center}
\begin{tabular}{c@{\qquad}ccc}
\toprule[0.8pt]
$K$ & Support F1 $\uparrow$ & Edit F1 $\uparrow$ & Value MAE $\downarrow$ \\
\midrule
0  & 0.263 & 0.016 & 73.820 \\
4  & 0.530 & 0.237 & 22.010 \\
16 & 0.750 & 0.466 & 14.303 \\
64 & \textbf{0.875} & \textbf{0.685} & \textbf{6.200} \\
\bottomrule[1.0pt]
\end{tabular}
\end{center}
\vspace{-10pt}
\end{table}

\subsection{Separating Naming Cues}
\label{sec:naming_results}

Schema and action names can be changed independently because task generation is separated from rendering. Their effect on TabWM is small once behavioral context is available.

\begin{table}[!ht]
\vspace{-10pt}
\footnotesize
\caption{Separating schema and action naming cues. Values are Edit F1.}
\label{tab:app_naming}
\setlength{\tabcolsep}{9pt}
\begin{center}
\begin{tabular}{llccc}
\toprule[0.8pt]
\multicolumn{2}{c}{Naming cues} & \multicolumn{3}{c}{Observed executions} \\
\cmidrule(lr){1-2}\cmidrule(lr){3-5}
Schema & Action & $K=4$ & $K=16$ & $K=64$ \\
\midrule
Readable & Readable & 0.194 & 0.498 & 0.684 \\
Opaque   & Readable & 0.178 & 0.494 & 0.681 \\
Readable & Opaque   & 0.195 & 0.498 & 0.682 \\
Opaque   & Opaque   & 0.188 & 0.487 & 0.674 \\
\bottomrule[1.0pt]
\end{tabular}
\end{center}
\vspace{-10pt}
\end{table}

With more observed executions, the four conditions converge to a narrow range. Together with the Qwen3 comparison in Table~\ref{tab:behavior_identification}, this separates semantic naming cues from the behavioral evidence TabWM uses for adaptation.

\subsection{Action-Specific Behavioral Evidence}
\label{sec:action_coverage}

The context-size sweep increases both the amount of behavioral evidence and the chance that the queried action has been observed before. We separate these effects by fixing the total context at $K=16$ and varying the number $m$ of demonstrations that execute the queried action. The remaining demonstrations use other actions from the same world. We evaluate the paired subset for which all values of $m\in\{0,1,2,4\}$ are available. As a direct retrieval control, an action-template baseline transfers the relational edit pattern observed for matching actions to the query.

\begin{table}[!ht]
\vspace{-10pt}
\footnotesize
\caption{Edit F1 at fixed $K=16$ as evidence for the queried action increases.}
\label{tab:action_coverage}
\setlength{\tabcolsep}{10pt}
\begin{center}
\begin{tabular}{c|cc}
\toprule[0.8pt]
Matching executions $m$ & TabWM & Template transfer \\
\midrule
0 & 0.048 & 0.003 \\
1 & 0.403 & 0.169 \\
2 & 0.599 & 0.188 \\
4 & \textbf{0.700} & 0.216 \\
\bottomrule[1.0pt]
\end{tabular}
\end{center}
\vspace{-10pt}
\end{table}

Direct evidence about the queried action is highly informative, but its benefit is not explained by replaying a previous consequence. The explicit template baseline improves little after the first matching execution, while TabWM continues to improve substantially as additional executions of the same action become available. TabWM therefore uses repeated executions to infer how an action's effect transfers across relational states rather than simply retrieving an observed edit pattern.

\subsection{Beyond the Pretraining Mechanism Prior}
\label{sec:ood_dynamics}

The shifts in Table~\ref{tab:world_changes} vary how mechanisms already represented during pretraining are assigned or combined. We next evaluate TabWM on four transition mechanisms introduced only at evaluation time. They retain the same relational state, action, and structured-edit interface, while their transition rules are absent from the \textsc{WorldRel} pretraining library.

Relational copy transfers a value across a foreign-key link. Cross-cell product makes an update depend multiplicatively on another cell. Redistribution spreads a quantity across related rows. Linked predicate gate conditions an effect on the state of a linked record. Table~\ref{tab:ood_mechanisms} separates effect localization from complete-edit prediction.

\begin{table}[!ht]
\vspace{-10pt}
\centering
\footnotesize
\caption{Behavioral transfer to evaluation-only transition mechanisms.}
\label{tab:ood_mechanisms}
\setlength{\tabcolsep}{7pt}
\begin{center}
\begin{tabular}{l|cc|c}
\toprule[0.8pt]
& \multicolumn{2}{c|}{Support F1}
& Edit F1 \\
Mechanism
& $K=0$ & $K=64$ & $K=64$ \\
\midrule
Relational copy
& 0.006 & 0.702 & 0.089 \\

Cross-cell product
& 0.376 & 0.910 & 0.031 \\

Redistribution
& 0.133 & 0.332 & 0.267 \\

Linked predicate gate
& 0.000 & 0.561 & 0.077 \\
\bottomrule[1.0pt]
\end{tabular}
\end{center}
\vspace{-10pt}
\end{table}

Behavioral context improves effect localization across all four mechanisms. Complete-edit prediction transfers less uniformly when the written value depends on a transformation not represented during pretraining. Redistribution is the strongest complete-edit case, consistent with its use of additive or multiplicative per-row arithmetic already supported by the numeric decoder. These results indicate that behavioral evidence transfers the relational footprint of an action farther than its exact quantitative transformation.

\subsection{Numeric Value Transfer Ablation}
\label{sec:numeric_reader_ablation}

The numeric decoder in \S\ref{sec:transition_prediction} combines a learned prediction with explicit additive and affine candidates. To isolate their contribution, we compare two numeric-decoder variants with the same initialization, training tasks, and optimization budget. The neural-only variant removes the explicit arithmetic candidates, while the full variant retains them.

\begin{table}[!ht]
\vspace{-10pt}
\centering
\footnotesize
\caption{Matched ablation of numeric value transfer. Values are Edit F1.}
\label{tab:numeric_reader_ablation}
\setlength{\tabcolsep}{10pt}
\begin{center}
\begin{tabular}{l|cc}
\toprule[0.8pt]
Variant & $K=16$ & $K=64$ \\
\midrule
Neural-only
& 0.380 & 0.476 \\

+ additive / affine
& \textbf{0.436} & \textbf{0.616} \\
\bottomrule[1.0pt]
\end{tabular}
\end{center}
\vspace{-10pt}
\end{table}

The arithmetic candidates improve complete-edit prediction, with a larger difference as more behavioral evidence becomes available. Support F1 is unchanged across the two variants. The explicit arithmetic paths therefore improve numeric value transfer without changing which relational edits are predicted.

\subsection{Sensitivity to Context Sampling and Numeric Tolerance}

\begin{figure}[!htbp]
\centering

\begin{tikzpicture}
\begin{axis}[
    width=0.46\textwidth,
    height=4.2cm,
    xlabel={Observed executions $K$},
    ylabel={Edit F1},
    xtick={4,16,64},
    ymin=0.15,
    ymax=0.75,
    axis lines=box,
    tick label style={font=\scriptsize},
    label style={font=\small},
    title style={font=\small},
    title={(a) Resampling behavioral context}
]
\addplot+[
    mark=*,
    line width=0.8pt,
    color=appblue,
    /pgfplots/error bars/y dir=both,
    /pgfplots/error bars/y explicit
]
coordinates {
    (4,0.222) +- (0,0.014)
    (16,0.469) +- (0,0.024)
    (64,0.680) +- (0,0.014)
};
\end{axis}
\end{tikzpicture}%
\hspace{0.02\linewidth}%
\begin{tikzpicture}
\begin{axis}[
    width=0.46\textwidth,
    height=4.2cm,
    xlabel={Relative numeric tolerance (\%)},
    ylabel={Edit F1},
    xtick={0,1,2,5,10},
    ymin=0.20,
    ymax=0.70,
    axis lines=box,
    tick label style={font=\scriptsize},
    label style={font=\small},
    title style={font=\small},
    title={(b) Numeric matching tolerance}
]
\addplot+[
    mark=*,
    line width=0.8pt,
    color=apporange
]
coordinates {
    (0,0.244)
    (1,0.469)
    (2,0.516)
    (5,0.581)
    (10,0.638)
};
\end{axis}
\end{tikzpicture}

\caption{Sensitivity of complete-edit prediction. (a) Mean Edit F1 across eight independently sampled tasks on the same worlds; error bars show the standard deviation across draws. (b) The same $K=16$ predictions rescored under increasingly permissive numeric tolerances.}
\label{fig:app_sensitivity}
\end{figure}
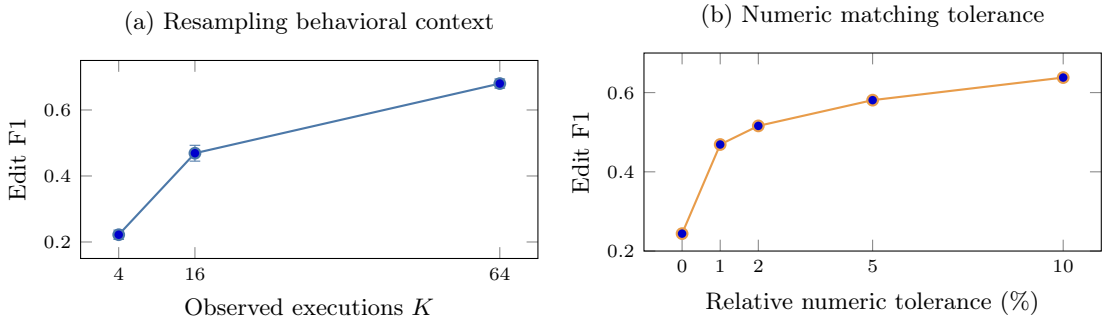

Two checks probe whether the main pattern depends on which executions are sampled or on the numeric matching threshold. Repeatedly sampling tasks from the same worlds produces little variation relative to the gain from additional behavioral context. Rescoring the same predictions under different numeric tolerances gives a smooth accuracy curve rather than a discontinuity around the default threshold. Figure~\ref{fig:app_sensitivity} summarizes both checks.

%% file: figures/prompt.tex
\begin{figure}[p]
\centering
\begin{adjustbox}{max totalsize={\linewidth}{0.90\textheight}}
\begin{minipage}{\linewidth}
\begin{promptbox}
\begin{Verbatim}[fontsize=\scriptsize, commandchars=\\\{\}]
You are a strict prediction assistant. Follow the instructions exactly.

# System
A relational database whose state is changed by actions. The rules governing
what an action changes are not given. Infer them from the demonstrations.

# Schema
  customers(id, stage, risk, quantity, score, units, stock, credit): 18 rows
  accounts(id, warehouses_id, customers_id, is_flagged, risk, tier): 11 rows
  warehouses(id, amount, priority, state, is_active, is_locked): 17 rows

Foreign keys:
  accounts.warehouses_id --> warehouses.id
  accounts.customers_id --> customers.id

Action signatures:
  restock_customer(target: customers.id, amount: number)
  reserve_warehouse(target: warehouses.id, mode: category)
  ...

# Task
Name: state-transition prediction

Description: Given an action and the records near it, predict the complete set
of cell-level edits that the action causes.

# Demonstrations

--- demonstration 1 ---
state before:
[customers]
  id=9, stage=stage_2, risk=39.4568, quantity=334, score=45.043, units=8, ...
[accounts]
  id=3, warehouses_id=2, customers_id=9, is_flagged=0, risk=120.856, tier=tier_0
[warehouses]
  id=2, amount=121, priority=terminated, state=queued, is_active=1, is_locked=1

action: restock_customer(target=9, amount=43)

edits:
  \{"op": "UPDATE_VALUE", "table": "customers", "row_id": 9,
   "column": "risk", "value": 5.40528\}

\textit{<demonstrations 2--3 omitted for display>}

--- demonstration 4 ---
state before:
[warehouses]
  id=14, amount=493, priority=new, state=queued, is_active=0, is_locked=0
[accounts]
  id=9, warehouses_id=14, customers_id=4, is_flagged=0, risk=492.989, tier=tier_0
[customers]
  id=4, stage=stage_2, risk=16.9865, quantity=0, score=27.294, units=11, ...

action: reserve_warehouse(target=14, mode=3)

edits:
  \{"op": "EMIT_EVENT", "table": "accounts"\}
  \{"op": "UPDATE_VALUE", "table": "warehouses", "row_id": 14,
   "column": "amount", "value": 496\}

# Query
state before:
[customers]
  id=8, stage=stage_0, risk=21.7258, quantity=197, score=13.9966, units=36, ...
[accounts]
  id=10, warehouses_id=15, customers_id=0, is_flagged=0, risk=247.983, tier=tier_1

action: reserve_customer(target=8, amount=9.7493)

# Output
Output JSON only, in exactly this form:
\{"edits": [\{"op": "...", "table": "...", "row_id": 0,
             "column": "...", "value": ...\}]\}

Valid operations are UPDATE_VALUE, UPDATE_STATE, EMIT_EVENT, and DELETE_ROW.
Use identifiers exactly as they appear in the task and output no text outside JSON.
\end{Verbatim}
\end{promptbox}
\end{minipage}
\end{adjustbox}
\caption{Abridged prompt template used in the readable naming condition for LLMs. Rows and two demonstrations are shortened only for display. The opaque condition preserves the same structure and values while consistently replacing schema, categorical, and action identifiers.}
\label{fig:qwen_prompt}
\end{figure}